\documentclass[sigconf]{acmart}
\usepackage{algorithm2e}
\usepackage{caption}
\usepackage{booktabs}
\usepackage{subcaption}
\usepackage{multirow}
\usepackage{lipsum}
\usepackage{float}
\usepackage{pgfplots}
\usepackage{pgfplotstable}
\usepackage{pgf-pie}
\usepackage{setspace}
\usepgfplotslibrary{fillbetween}
\usepgfplotslibrary{external}
\pgfplotsset{width=7cm,compat=1.18}
\usepackage{thmtools, thm-restate}
\usepackage{enumitem}
\usepackage{lipsum}
\usepackage{adjustbox}
\usetikzlibrary{patterns}
\newcommand{\camcomment}[1]{#1}
\usepackage{pifont}
\AtBeginDocument{%
  }

\copyrightyear{2026}
\acmYear{2026}
\setcopyright{cc}
\setcctype{by}
\acmConference[WWW '26] {Proceedings of the ACM Web Conference 2026}{April 13--17, 2026}{Dubai, United Arab Emirates.}
\acmBooktitle{Proceedings of the ACM Web Conference 2026 (WWW '26), April 13--17, 2026, Dubai, United Arab Emirates}
\acmISBN{979-8-4007-2307-0/2026/04}
\acmDOI{10.1145/XXXXXX.XXXXXX}


\begin{document}

\title{Efficient Content-based Recommendation Model Training via Noise-aware Coreset Selection}

\author{Hung Vinh Tran}
\email{h.v.tran@uq.edu.au}
\affiliation{%
  \institution{The University of Queensland}
  \city{Brisbane}
  \state{Queensland}
  \country{Australia}
  \postcode{4072}
}
\author{Tong Chen}
\email{tong.chen@uq.edu.au}
\affiliation{%
  \institution{The University of Queensland}
  \city{Brisbane}
  \state{Queensland}
  \country{Australia}
  \postcode{4072}
}

\author{Hechuan Wen}
\email{h.wen@uq.edu.au}
\affiliation{%
  \institution{The University of Queensland}
  \city{Brisbane}
  \state{Queensland}
  \country{Australia}
  \postcode{4072}
}

\author{Quoc Viet Hung Nguyen}
\email{henry.nguyen@griffith.edu.au}
\affiliation{%
  \institution{Griffith University}
  \city{Gold Coast}
  \state{Queensland}
  \country{Australia}
  \postcode{4222}
}

\author{Bin Cui}
\email{bin.cui@pku.edu.cn}
\affiliation{%
  \institution{Peking University}
  \city{Beijing}
  \country{China}
}

\author{Hongzhi Yin}
\email{h.yin1@uq.edu.au}
\authornote{Corresponding author.}
\affiliation{%
  \institution{The University of Queensland}
  \city{Brisbane}
  \state{Queensland}
  \country{Australia}
  \postcode{4072}
}

\renewcommand{\shortauthors}{Tran et al.}

\begin{abstract}
Content-based recommendation systems (CRSs) utilize content features to predict user-item interactions, serving as essential tools for helping users navigate information-rich web services. However, ensuring the effectiveness of CRSs requires large-scale and even continuous model training to accommodate diverse user preferences, resulting in significant computational costs and resource demands. A promising approach to this challenge is coreset selection, which identifies a small but representative subset of data samples that preserves model quality while reducing training overhead. Yet, the selected coreset is vulnerable to the pervasive noise in user-item interactions, particularly when it is minimally sized. To this end, we propose Noise-aware Coreset Selection (NaCS), a specialized framework for CRSs. NaCS constructs coresets through submodular optimization based on training gradients, while simultaneously correcting noisy labels using a progressively trained model. Meanwhile, we refine the selected coreset by filtering out low-confidence samples through uncertainty quantification, thereby avoid training with unreliable interactions. Through extensive experiments, we show that NaCS produces higher-quality coresets for CRSs while achieving better efficiency than existing coreset selection techniques. Notably, NaCS recovers 93-95\% of full-dataset training performance using merely 1\% of the training data.
The source code is available at \href{https://github.com/chenxing1999/nacs}{https://github.com/chenxing1999/nacs}.
    
\end{abstract}




\maketitle


\section{Introduction}
With the advent of the information explosion, deep recommender systems have become essential in navigating daily life. 
These systems are the cores of various online services, such as e-commerce websites \cite{pfadler2020billion} 
, social networks \cite{naumov2019deep} 
, and streaming platforms \cite{steck2021deep}. 
Meanwhile, with the increasing diversity of content information available about both users and items, there is a shift from traditional collaborative filtering to content-based recommendation systems (CRSs). CRSs leverage side features such as user genders and movie genres for predictions on user-item interactions to facilitate more accurate recommendations \cite{mixfm2024tpami}, where click-through rate (CTR) prediction is a typical task \cite{barsctr2021,deepfm2017,mao2023finalmlp}. 

However, to fully capture diverse user interests, CRS models commonly default to training with large-scale datasets \cite{hung2017computing,yin2016spatio}. On the one hand, the massive training dataset, which can involve billions of user-item interaction logs (e.g., the Google Play dataset \cite{cheng2016wide}), has placed a heavy burden on time cost, computational resources, and carbon footprint \cite{sachdeva2022infinite, svpcf2022}. On the other hand, in a production environment, a recommendation model requires frequent updates with recent data to stay relevant. 
As per \cite{lee2023periodic_update}, a 15\% model performance degradation is witnessed if model updates are delayed by a day. Consequently, the efficiency hurdle is amplified by recurrently training CRS models. 
In addition, in the recent surge of deploying large language models (LLMs) as CRS backbones \cite{liu2024once,yuan2024fellas,kieu2025enhancing}, 
the time and resource overheads are a major bottleneck for training LLM-based recommenders at scale. 
As such, these practicality obstacles lead to a strong demand to reduce the computational costs of training CRS models. 

In this regard, a natural solution to the efficiency bottleneck is to identify a small, representative subset of the data, known as a \textit{coreset}, that serves as a proxy of the full dataset.
For recommender systems, most data-centric research regarding training efficiency still focuses on classic collaborative filtering tasks \cite{svpcf2022,sachdeva2022infinite,qu2024sparser}, which are non-trivial to generalize to content-based recommendation due to the higher level of complexity introduced by user and item feature interactions. 
Recently, some work \cite{wang2023gradient,qin2025d2k} has started exploring data-centric efficient training for CRSs. 
For example, CGM \cite{wang2023gradient} aims to generate a condensed synthetic dataset via gradient matching to replicate the model's training trajectory with the full dataset. However, to facilitate continuous gradient-based optimization, CGM has to disobey the sparse nature of categorical input features\footnote{That is, only a few values of one feature field are activated \cite{fm2010}, e.g., the one-hot encoding for user occupation. Note that CRSs predominantly use categorical features, where numerical features are converted into categorical ones via bucketing \cite{bars2022}.}, and resort to dense input features. Such dense features are hardly representative of reality -- one can think of user occupation and product origin country as two examples, where it is impossible to activate all possible values in a single sample. Consequently, the performance of the trained models is suboptimal. Furthermore, the reliance on dense data representations leads to unnecessary computational and storage overheads, counteracting the benefit of reducing the training sample size. 
More recently, D2K \cite{qin2025d2k} provides a data-centric approach for CRSs. 
However, its focus is on retrieving informative historical samples to better adapt to unseen inputs during inference, where the training process still requires the full dataset.  
Thus, selecting coresets from raw data for efficient model training remains largely underexplored in CRSs.



This leads to the first challenge of \textit{efficiently selecting the most representative samples from a large-scale dataset.}
To ensure practicality, the coreset selection process itself must be significantly more efficient than training on the full dataset. This requires developing strategies that can scale and avoid costly computations like repetitive full-batch gradient evaluations \cite{wang2023gradient}, while warranting maximum utility for CRS model training. In the meantime, the second challenge arises from \textit{the inherent noise within the implicit user feedback in CRSs}. Unlike explicit feedback that carries users' fine-grained preferences via ratings or reviews, implicit feedback like clicks and purchases only provides indirect signals about user preferences, yet is the main label type used in CRSs (e.g., user clicks in CTR prediction tasks). This presents a fundamental issue because implicit feedback does not necessarily reflect the true user intention -- for example, a user might click on an item for reasons unrelated to preference, such as curiosity, popularity, or by mistake \cite{wang2021clicks}, while a lack of interaction does not always indicate disinterest \cite{ye2022don}. This ambiguity introduces noise and bias into the observed data, rendering most general coreset selection methods \cite{mirzasoleiman2020coresets} unsuitable for CRSs due to their noise-free assumptions. This is because a coreset is selected by maximizing its capability to replicate a model's training behavior (e.g., gradients or interim accuracy) on the full data, which can be easily misled by the noisy preference labels in CRSs. As a result, the constructed coreset is prone to containing incorrect supervision signals, and risks distorting model training and degrading its recommendation performance.

\begin{figure*}
    \centering
    \includegraphics[width=0.75\linewidth]{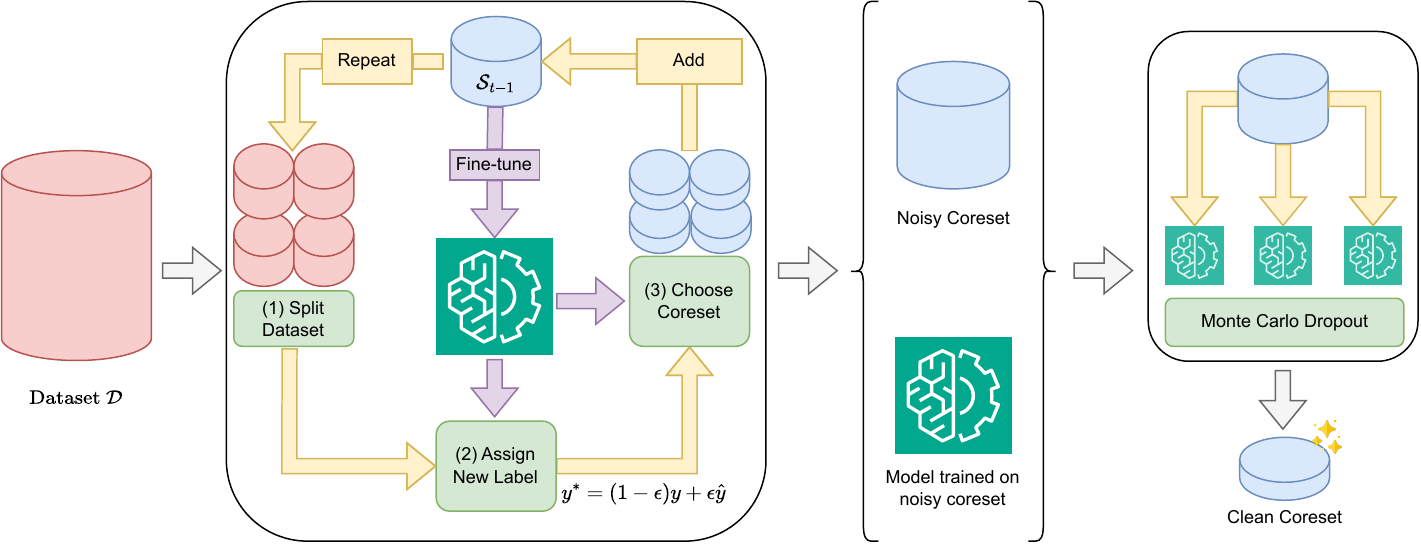}
    \caption{\textbf{The overview of NaCS.} First, we progressively construct the coreset while refining the labels. Next, we apply Monte Carlo Dropout to estimate uncertainty and remove highly uncertain samples, resulting in the final clean coreset.}
    \label{fig:pipeline}
\end{figure*}

Motivated by these challenges, we propose \textbf{\underline{N}oise-\underline{a}ware \underline{C}oreset \underline{S}election (NaCS}), an efficient and accurate coreset selection framework for content-based recommendation. Specifically, to tackle the first challenge regarding efficiency, we frame the coreset selection as a submodular optimization problem \cite{mirzasoleiman2020coresets,yang2023crest,dang2025colm}, which offers a theoretical framework for set optimization problems. 
In a nutshell, our submodular optimization objective aims to choose a representative subset that maximizes a utility function under budget constraints. For the utility function, a common choice is the similarity between model gradients w.r.t. the full data and coreset \cite{mirzasoleiman2020coresets}, yet it incurs heavy computations associated with calculating the derivatives of all model layers. In NaCS, instead of backpropagating the whole model, only the final layer gradients are utilized to efficiently capture the gradient variance. Furthermore, to achieve a linear time complexity w.r.t. the full data size, we solve the submodular optimization objective via a stochastic greedy approach \cite{mirzasoleiman2015lazier}, thus accelerating the selection process while retaining approximation guarantees. Additionally, we process the stochastic greedy algorithm in concurrent batches, which takes full advantage of GPU parallelization and significantly reduces the execution time compared to outgoing open source libraries \cite{kaushal2022submodlib,guo2022deepcore}.

To tackle the second challenge of noisy user-item interactions, we propose a two-stage data denoising paradigm. 
The first step denoises the label through a progressive label self-correction approach \cite{wang2021proselflc,wang2022proselflc}, where we utilize the model trained on the coreset to gradually assign a new label for each sample. 
Specifically, starting from a randomly initialized model, we iteratively select samples whose gradient best mimics the whole dataset gradient estimated by the current model. In each iteration of the first stage, the model trained on the current coreset is used to refine the labels of the whole dataset. As the coreset grows and the model becomes more accurate, the corrected labels become correspondingly more reliable, allowing us to mitigate the impact from noisy labels to coreset our selection.
After obtaining the initial coreset, the second stage further refines it by estimating each sample's quality via uncertainty estimation \cite{wang2021wsdm_denoise}. For this purpose, we employ the Monte Carlo Dropout \cite{gal2016dropout}, which involves performing multiple forward passes through the network with dropout layers enabled at test time. This process generates a distribution of predictions for each sample, allowing us to quantify sample-specific uncertainty. Then, samples with the highest uncertainty are removed, protecting the recommendation model from outliers in the small coreset.

To demonstrate our method's effectiveness, we conduct extensive experiments on three representative large-scale datasets for content-based recommendation, namely Criteo \cite{Criteo}, Avazu \cite{Avazu}, and KDD \cite{KDD}. The experiment results show that we can recover up to 93-95\% of model performance on three datasets using only 1\% of the original dataset size. Additionally, we further verify the versatility of NaCS in a text-based recommendation scenario. Our method showed strong generalization, maintaining competitive performance while significantly reducing data and computational resource requirements. 
We summarize our contributions as follows:
\begin{enumerate}
    \item We highlight the training efficiency issue from a data-centric perspective and point out the necessity of denoising implicit feedback in coreset selection, which is an essential but overlooked challenge in content-based recommendation systems.
    \item We design a novel framework, namely NaCS, to select a clean coreset efficiently. Firstly, we employ a submodular optimization and a progressive self-correction approach to gradually build the coreset. Secondly, we apply Monte Carlo dropout to further denoise the coreset from the first step, producing a more compact one.
    \item Extensive experiments on large-scale datasets verify NaCS's performance and efficiency. The results show that our proposed approach not only can produce a robust coreset efficiently, but also demonstrates competitive performance in text-based recommendation.
\end{enumerate}


\section{Related Work}
\label{sec:related}
This section summarizes the related work, while we also provide a more extensive discussion in Appendix \ref{sec:extended-rw}.

\noindent \textbf{Coreset Selection}: Various heuristics for coreset selection train full \cite{birodkar2019semantic} or proxy models \cite{coleman2019selection} to assign scores \cite{coleman2019selection, paul2021deep, tonevaempirical2019,xiao2024feature}, but they require training on the entire dataset and lack a trade-off between performance and efficiency. 
Following previous work \cite{yang2023crest,dang2025colm,mirzasoleiman2020coresets}, our method adopts submodular optimization, with two key innovations: progressive label correction and a denoise module. Moreover, we implement an efficient stochastic greedy optimization algorithm, enhancing its real-world practicality for CRS training.

\noindent \textbf{Dataset Reduction for Recommender Systems}: Recently, due to the high cost of training for recommender systems, various dataset reduction methods have been proposed for recommender systems, including both dataset distillation \cite{zhang2025td3,sachdeva2022infinite,wang2023gradient,gao2025relational} and coreset selection \cite{mei2025goracs,lin2024data}. However, they are working on collaborative filtering \cite{sachdeva2022infinite}, sequential recommendation \cite{zhang2025td3}, or LLM-based recommendation tasks \cite{mei2025goracs,lin2024data}. CGM \cite{wang2023gradient}, which was proposed for CRs, employs gradient matching and dense representation, leading to high computational costs.

\noindent \textbf{Implicit Feedback Denoising}: Recent studies \cite{wang2021wsdm_denoise} have highlighted the substantial noise in implicit feedback datasets due to their characteristics.
To tackle this, initial studies leverage auxiliary information (e.g. ``dwell time'' \cite{kim2014modeling}, ``like'' \cite{bian2021denoising}), which can be costly to collect. Wang et al. \cite{wang2021wsdm_denoise} discovered that the noisy samples can be found through the loss values. Although some methods \cite{he2024dcf,zhang2025shapley} have been proposed to denoise implicit feedback, they still have an accuracy-oriented focus, with none or limited reduction on the training data size. In contrast, we propose a complementary design between CRS data denoising and coreset selection.

\vspace{-0.3cm}
\section{Preliminaries}
\label{sec:prelim}
In this section, we first introduce the key concepts of content-based recommendation by defining a classic task, namely click-through rate (CTR) prediction. Then, we provide the background for coreset selection approaches, submodular optimization problems, and their solutions -- the core theoretical framework for our approach. 

\vspace{-0.3cm}
\subsection{Content-based Recommendation}
In CTR prediction, a dataset $\mathcal{D}$ is a set of samples $(\mathbf{x}, y)$, where $\mathbf{x}$ is a multi-dimensional vector combining both user, item and context features, and $y$ is a binary label indicating positive ($y=1$, clicked) or negative ($y=0$, not clicked) user interaction with the item. 
For $\mathbf{x}$, its features are commonly a collection of sparse binary features from multiple fields (i.e., user occupation and product category) \cite{marcuzzo2022recommendation}, called categorical features. 
These categorical features are typically embedded into dense vectors using randomly initialized embedding tables. 
This preprocessing facilitates downstream learning by converting heterogeneous raw inputs into a unified, low-dimensional representation. 
These representations are then fed into a neural network, commonly called model head, to capture feature interactions and predict user engagement signals, represented by the model prediction $\hat{y}$.
\camcomment{Figure \ref{fig:sample-process} shows an example pipeline of CTR prediction model.}
The whole model will be trained with log loss:
\begin{equation}
\label{eq:logloss}
    L = -y \ln (\hat{y}) - (1 - y) \ln (1 - \hat{y}),
\end{equation}
which computes the difference between $y$ and $\hat{y}$.

\camcomment{
\begin{figure}
    \centering
    \includegraphics[width=0.85\linewidth]{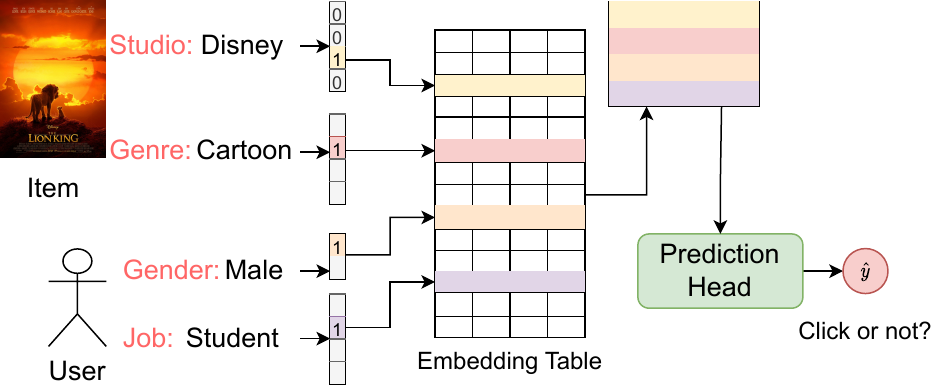}
    \caption{Example process of CRSs.}
    \label{fig:sample-process}
    \vspace{-0.3cm}
\end{figure}
}

\subsection{Coreset Selection}

Existing coreset methods \cite{mirzasoleiman2020coresets,yang2023crest} search for a weighted subset of examples that match the full dataset training gradient.
Formally, the target is to find the smallest subset $\mathcal{S} \subseteq \mathcal{D}$ and corresponding weight $\gamma_j$ for each element that approximate the full gradient with an error at most $\epsilon > 0$ for all possible values of model weight $\omega_t$ in the hypothesis space $\mathcal{W}$:
\begin{equation}
\label{eq:ori-opt}
    \mathcal{S}^* = \arg\min_{\mathcal{S} \subseteq \mathcal{D}} \lvert S \rvert, 
    \text{ s.t. } 
    \max_{\omega_t \in \mathcal{W}} \lVert \sum_{i \in \mathcal{D}} g_{i,t} - \sum_{j \in \mathcal{S}} \gamma_j g_{j,t} \rVert  \leq \epsilon,
\end{equation}
where $\lvert \cdot \rvert$ is the cardinality, $\lVert \cdot \rVert$ is the $L^2$ norm, and $g_{i,t}$ is the gradient respected to the model weight $\omega_t$ for $i$-th sample.
The above problem requires us to calculate through all possible gradients in the hypothesis space, which is infeasible. 
For neural networks, it is shown that the final layer gradient $g^L_{\cdot,\cdot}$ can well capture the variation of gradient norm \cite{katharopoulos2018not,mirzasoleiman2020coresets}, which we further empirically verify in Appendix \ref{sec:last-layer}. Thus, we can rewrite the constraint in Eq. \ref{eq:ori-opt} as:
\begin{equation}
    C - \sum_{i \in \mathcal{D}} \min_{j \in \mathcal{S}} \left \Vert g^L_{i,t} - g^L_{j,t} \right \Vert \leq \epsilon,
\end{equation}
where $C$ is a large constant.
As we are interested in finding the best subset for a given training budget $k$, we can solve the following objective, which is the dual problem w.r.t. the above one:
\begin{equation}
    \mathcal{S}^*_t = \arg \max_{\mathcal{S} \subseteq \mathcal{D}} \left( C - \sum_{i \in \mathcal{D}} \min_{j \in \mathcal{S}} \left \Vert g^L_{i,t} - g^L_{j,t} \right \Vert \right),
    \text{ s.t. }
    \lvert S \rvert \leq k.
    \label{eq:opt-obj-1}
\end{equation}

\subsection{Submodular Optimization}
Submodular optimization is a powerful framework in combinatorial optimization characterized by the diminishing returns property.
Formally, given a function $F: 2^{\mathcal{D}} \rightarrow \mathbb{R}$, $F$ is submodular if:
\begin{equation}
    \forall \mathcal{S} \subseteq \mathcal{T}, i \notin \mathcal{T}, F(\mathcal{S} \cup i) - F(\mathcal{S}) \geq F(\mathcal{T} \cup i) - F(\mathcal{T}).
\end{equation}
The optimization objective in Eq. \ref{eq:opt-obj-1} is a constrained monotone\footnote{$F$ is monotone if $\forall \mathcal{S} \subset \mathcal{T}, F(\mathcal{S}) < F(\mathcal{T})$.} submodular function. 
We leverage this structure to develop an efficient, theoretically grounded selection strategy that enables data summarization under the computational and storage constraints of large-scale machine learning.

Maximizing a monotone submodular function under a cardinality constraint is NP-hard, so we rely on approximation algorithms rather than exact solutions. 
A widely used approach is the greedy algorithm: starting from an empty set, we iteratively add the element that yields the maximum marginal gain in the objective until reaching the size limit. 
Due to the diminishing returns property of submodular functions, the greedy algorithm provably achieves the optimal approximation ratio of $1 - 1/e$ for monotone submodular maximization under a cardinality constraint \cite{Nemhauser1978}. 
Most existing coreset methods \cite{dang2025colm,yang2023crest,mirzasoleiman2020coresets} apply this standard greedy algorithm to approximate their objective.
However, the above naive greedy algorithm has the time complexity of $O(\vert \mathcal{D} \vert \times k)$, which is not scalable for large datasets. 
So in this work, we adopt the stochastic greedy algorithm \cite{mirzasoleiman2015lazier} due to the significantly larger scale of CTR datasets compared to those in computer vision.

\section{Methodology}
\label{sec:method}
As shown in Figure \ref{fig:pipeline}, our method consists of three main components. First, we introduce a coreset selection method that employs submodular optimization to efficiently search for key samples. 
Second, to better model user preference, we relax the discrete binary label into a continuous label through a progressive self correction manner. 
Third, by building an uncertainty quantification scheme upon Monte Carlo Dropout, we further identify and remove highly uncertain samples from the coreset.

\subsection{CRS Coreset Selection}

Given that we use Adam optimizer,  we adapt the submodular coreset formulation introduced in \cite{dang2025colm} to our setting. Firstly, let's revisit the Adam optimization process.
Adam is an optimization algorithm designed to adaptively scale the gradient for each parameter, enabling efficient training of deep neural networks. It combines the advantages of first- and second-order estimation to compute adaptive learning rates for each parameter. At each timestep $t$, Adam updates parameters using biased estimates of the first moment $m_t$ and the second moment $v_t$ of the gradients $g_t = \nabla_\theta L(\theta_t)$. These are computed as $m_t = \beta_1 m_{t-1} + (1 - \beta_1) g_t$ and $v_t = \beta_2 v_{t-1} + (1 - \beta_2) g_t \odot g_t$, where $\beta_1, \beta_2 \in [0,1)$ are decay rates, and $\odot$ is the element-wise product. To correct the bias introduced in the initial steps, Adam uses bias-corrected estimates: $\hat{m}_t = m_t / (1 - \beta_1^t)$ and $\hat{v}_t = v_t / (1 - \beta_2^t)$. The parameter update rule is then given by:
\[
\theta_{t+1} = \theta_t - \alpha \frac{\hat{m}_t}{\sqrt{\hat{v}_t} + \varepsilon},
\]
where $\alpha$ is the learning rate and $\varepsilon$ is a small constant. 

To construct a coreset that closely approximates the training dynamics of the full dataset, we focus on preserving the estimated gradients. As mentioned, in practice, we only preserve the final layer gradient to reduce computational cost, and it has been shown to be sufficient \cite{mirzasoleiman2020coresets}. 
Specifically, we aim to select a subset $\mathcal{S}^* \subseteq \mathcal{D}$ that maximize the following objective:
\begin{equation}
    \label{eq:opt-obj}
        \mathcal{S}^* = \arg\max_{S \subseteq \mathcal{D}} \left( C - \sum_{i \in \mathcal{D}} \min_{j \in \mathcal{S}} \left \Vert \hat{g}^L_{i,t} - \hat{g}^L_{j,t} \right \Vert \right)
     \text{ s.t. } \vert \mathcal{S}^* \vert \leq k,
\end{equation}
where $C$ is a constant, $\hat{g}^L_{i,t}$ is the estimated gradient of final layer for $i$-th sample at timestep $t$, and $k$ is the coreset budget. Formally, we compute $\hat{g}_{i,t}$ as following:
\begin{align}
    \hat{g}_{i,t} &= \hat{m}_{i,t} / ( {\sqrt{\hat{v}_{i,t}} + \varepsilon} ), \\
    \hat{m}_{i,t} &= \left( \beta_1 m_{t -1} + (1 - \beta_1) g_{i, t} \right) / \left( {1 - \beta_1^t} \right), \\
    \hat{v}_{i,t} &= \left( \beta_2 v_{t - 1} + (1 - \beta_2) g_{i, t} \odot g_{i, t} \right) / \left( {1 - \beta_2^t} \right),
\end{align}
where $\hat{m}_{i,t}$ and $\hat{v}_{i,t}$ are the first- and second-order estimation of gradient in the current timestep $t$ when adding the $i$-th sample gradient.  
The optimization objective in Eq.~\ref{eq:opt-obj} is monotone submodular. 
To further reduce computational complexity, we adopt the common practice of not constructing a single coreset for the entire dataset \cite{dang2025colm,yang2023crest,mirzasoleiman2020coresets}. Instead, we partition the dataset $\mathcal{D}$ into smaller disjoint batches $ \{ \mathcal{D}_1, \mathcal{D}_2, \dots, \mathcal{D}_B\} $ and identify a compact coreset $S^*_b$ within each batch:
\begin{equation}
\label{eq:opt-obj-batch}
        \mathcal{S}_b^* = \arg\max_{S \subseteq \mathcal{D}_b} \left( C - \sum_{i \in \mathcal{D}_b} \min_{j \in \mathcal{S}} \left \Vert \hat{g}^L_{i,t} - \hat{g}^L_{j,t} \right \Vert \right),
     \text{ s.t. } \vert \mathcal{S}_b^* \vert \leq \frac{k}{B}.
\end{equation}
We implement the stochastic greedy algorithm in PyTorch, leveraging its capabilities for GPU acceleration.
A significant aspect in our approach is the batch-wise execution of the algorithm on the GPU, allowing for the simultaneous solution of multiple submodular optimization problems in Eq. \ref{eq:opt-obj-batch}. This parallel processing capability significantly enhances computational efficiency, leading to substantially faster performance compared to previously established methods such as those employed in previous papers \cite{yang2023crest,mirzasoleiman2020coresets} and other available open sources \cite{kaushal2022submodlib,guo2022deepcore}. 

Since the optimization objective in Eq.~\ref{eq:opt-obj} depends on the model parameters, a practical approach is to identify multiple smaller subsets incrementally throughout the training process, rather than searching for a single global coreset upfront. Concretely, we first initialize a random model $f$, then we iteratively select subsets in $N_{choose}$ steps. For each step, based on the current model parameters and gradient estimation, we identify the optimal subset $\mathcal{S}$ of size $k / N_{choose}$. We fine-tune the model $f$ on the updated subset. The process is repeated for $N_{choose}$ times, allowing the coreset to adapt to various training stages of the model. 

\subsection{Progressive Label Self-Correction}

Implicit feedback datasets are inherently noisy since they capture user behaviors (e.g., clicks) rather than true preferences. For instance, a user might click on an item out of curiosity, due to an attractive thumbnail, or by mistake -- none of which necessarily indicates genuine interest. This noise becomes especially problematic in small datasets, where the gap between behavior and actual preference is harder to bridge. 
To this end, we employ progressive self-label correction \cite{wang2021proselflc,wang2022proselflc} to approximate the true user preference. On a high level, this method gradually approximates the true user preference $y^*$ through the current model prediction $y$, based on the following equation:
\begin{equation}
    y^* = (1 - \epsilon) y + \epsilon\hat{y},
\end{equation}
where $y^*$ is the new label, $y$ is the annotated label, $\hat{y}$ is the model prediction, and $\epsilon$ is a value that changes over time based on the current training timestep and model prediction. 
We will describe how to compute $\epsilon$, which is the combination of ``global'' and ``local'' trust, in the following part. We calculate global trust $g(t)$, which is shared between all samples, as following: 
\begin{equation}
\begin{split}
    g(t) &= h(t / T - 0.5, \beta), 
    \\
    h(\alpha, \beta) &= \frac{1}{1+\exp(-\alpha \times \beta)},
\end{split}
\end{equation}
where $t$ is the current timestep, and $T$ is the total number of time steps, $\beta$ is a hyperparameter, which we empirically set to 16 as per \cite{wang2021proselflc}. The global trust $g(t)$ increases over time, reflecting the growing confidence in the model's predictions as its performance improves.
Next, for each sample, we define the local trust $l(\hat{y})$, which is different between each input sample, based on the model's confidence in its output. This is calculated as:
\begin{equation}
    l(\hat{y}) = 1 - H(\hat{y}) / H(u),
\end{equation}
where $H(\cdot)$ is the entropy of sample, $u$ is the uniform distribution. With the above definition, as the model has more confidence in its output $\hat{y}$ (lower entropy), the local trust will further increase. When the model has less confidence in its output, the entropy approaches the uniform distribution's entropy, leading to lower local trust. Finally, we can calculate $\epsilon$ as following:
\begin{equation}
    \epsilon = g(t) \times p(\hat{y}).
\end{equation}
Appendix \ref{sec:learn-curve} depicts the progression of $\epsilon$ over training epochs.
Last but not least, we consider $y^* \in \left[ 0, 1 \right]$ as our new soft label and calculate the loss function as follows:
\begin{equation}
    \label{eq:proselflc-main}
    L_{lc}(\hat{y}, y) = -y^* \ln(\hat{y}) - (1-y^*)\ln(1 - \hat{y}).
\end{equation}

As this method requires minimal overhead, it can be done efficiently at each training iteration $t$ on the whole dataset level. This denoising approach is computationally efficient compared to prior methods \cite{he2024dcf}, incurring minimal overhead. 
\camcomment{For example, BOD \cite{wang2023efficient} relies on bi-level gradient optimization, whereas our method requires only a single forward pass to compute the updated label.}

\RestyleAlgo{ruled}
\LinesNumbered
\begin{algorithm}[th]
\setstretch{0.8} 
\caption{Noise-aware Coreset Selection for Content-based Recommendation Models}
\label{alg:main}
\SetKwRepeat{Do}{do}{while}
\SetKwComment{Comment}{/* }{ */}

\KwData{A training dataset $\mathcal{D}$, subset size $k$, number of subset training steps $N_{choose}$}
\KwResult{Subset $\mathcal{S}$}

\Comment{Step 1. Select Coreset}
$\mathcal{S} \gets \emptyset$, $\mathbf{m}_{grad} \gets \mathbf{0}, \mathbf{v}_{grad} \gets \mathbf{0}$\;
Initialize a random model $f$\;
\For{Epoch $e = 1,2 \dots N_{choose} $ }{
    \For{Batch $b=1,2, \dots, B_{choose}$}{
        $\mathcal{D}_b \gets$  Sample from $\mathcal{D}$\;
        $\mathcal{S}_b \gets $ Solve the optimization objective in Eq.~\ref{eq:opt-obj-batch} with gradient from $L_{lc}$\;
        $\mathcal{S} \gets \mathcal{S} \cup \mathcal{S}_b$\;
    }
    Train $f$ on $\mathcal{S}$ using loss $L_{lc}$ (Eq.~\ref{eq:proselflc-main})\;
    Update $\mathbf{m}_{grad}$ and $\mathbf{v}_{grad}$ accordingly\;
}

\Comment{Step 2. Denoise Coreset}
$\mathbf{m}_{loss} \gets \mathbf{0}, \mathbf{v}_{loss} \gets \mathbf{0}$\;
Enable dropout in $f$\;
\For{$i = 1,2, \dots, N_{denoise}$}{
    Update $\mathbf{m}_{loss}$ and $\mathbf{v}_{loss}$ according to the loss values with Monte Carlo Dropout on $f$\;
}
$\mathcal{S} \gets \text{Top } 90\% $ of upper bound loss (Eq. \ref{eq:upper-loss})\;
\vspace{-.1cm}
\end{algorithm}

\subsection{Coreset Denoising}

\begin{figure}
    \centering
    \pgfplotstableread[col sep=comma]{figures/data/loss_avazu_subset_deepfm.csv}\avazu
\pgfplotstableread[col sep=comma]{figures/data/loss_criteo_subset_deepfm.csv}\criteo

\begin{tikzpicture}
    \begin{axis}[
        width=0.98\linewidth,
        height=3.5cm,
        xlabel={Sample ID},
        ylabel={Log Loss},
        legend style={anchor=north,at={(0.2,0.95)},legend columns=1},
    ]
    \addplot+[mark size=1pt] table[
        x=idx, 
        y=values
    ]
    {\criteo};

    \addplot+[mark size=1pt] table[
        x=idx, 
        y=values
    ]
    {\avazu};
    \legend{Criteo, Avazu}
    \end{axis}
    
\end{tikzpicture}
    \vspace{-0.3cm}
    \caption{Log loss of trained DeepFM model on 1\% coreset. \textmd{``Sample ID'' is the index in the sorted array based on log loss.}}
    \label{fig:sorted-loss}
    \vspace{-0.4cm}
\end{figure}

Despite the previous label denoise and coreset selection step, the dataset still contains noise.  
Although we do not have access to users' true preferences, we can identify the noisy samples through loss values \cite{wang2021wsdm_denoise}. 
This can be shown empirically in Figure \ref{fig:sorted-loss}.
As our coreset is substantially smaller than the original dataset, the noisy samples have a much larger negative impact on model performance. 

To this end, we employ Monte Carlo Dropout \cite{gal2016dropout} to estimate per-sample noise: by performing multiple stochastic forward passes with dropout, we compute an upper bound on the loss for each sample, using the following equation:
\begin{equation}
\label{eq:upper-loss}
    L_{upper} = m_{loss} + 1.96 \sqrt{\frac{v_{loss}}{N_{denoise}}},
\end{equation}
where $m_{loss}$ and $v_{loss}$ are the first- and second-order estimation for loss value approximated with $N_{denoise}$, which we set to $10$, samples for Monte Carlo approximation; the constant 1.96 corresponds to a 95\% confidence interval.
We remove $10\%$ of samples with the highest estimated loss in our main experiments, while further hyperparameter studies are available in Appendix \ref{sec:exp-hyperparam}.
Intuitively, if the model exhibits high variance in its predicted loss, it suggests uncertainty or potential noise in that sample.
By employing the upper confidence bound on the loss values, we effectively prioritize samples with both high mean loss and high uncertainty.
This method serves as a simple denoising strategy, which is demonstrated to have a competitive performance. 
\camcomment{We leave a deeper exploration of its potential for future work.}
%
The pseudo code for our framework is provided in Algorithm \ref{alg:main}.

\subsection{Efficient Submodular Optimization Implementation}

\begin{table}[t]
    \centering
 \caption{\textbf{Stochastic greedy algorithm implementation comparison.} \textmd{The results are shown in seconds. $k$ is number of point to select, $\vert \mathcal{D} \vert$ is the dataset size, and $\epsilon$ denotes the error rate for stochastic greedy algorithm. The second header of Submodlib denotes the similarity kernel implementation. ``-'' denotes OOM. ``B'' refers to our batch size. Best result in each line is marked in bold.}}
    \renewcommand{\arraystretch}{0.75}
    \vspace{-0.3cm}
        \begin{tabular}{lccccc}
            \toprule
            (k, $\epsilon$, $\vert \mathcal{D} \vert$)  & Deep & \multicolumn{2}{c}{Submodlib} & \multicolumn{2}{c}{Ours} \\
            \cmidrule(lr){3-4} \cmidrule(lr){5-6}
            & Core & CPU & GPU & B=1 & B=4 \\
            \midrule
            (50, $10^{-3}$, $10^2$) & 0.0028 & \textbf{0.0015} & 0.0128 & 0.0301 & 0.0073 \\
            (50, $10^{-3}$, $10^3$) & 0.1610 & 0.0853 & 0.0244 & 0.0316 & \textbf{0.0078} \\
            (50, $10^{-3}$, $10^4$) & - & 10.950 & 5.0140 & 0.0591 & \textbf{0.0341} \\
            (100, $10^{-3}$, $10^4$) & - & 10.944 & 4.9948 & 0.0676 & \textbf{0.0364} \\
            (100, $10^{-4}$, $10^4$) & - & 12.5461 & 6.5867 & 0.0742 & \textbf{0.0434} \\
            (200, $10^{-4}$, $10^4$) & - & 12.6004 & 6.6277 & 0.0893 & \textbf{0.0478} \\
            \bottomrule
    \end{tabular}
        \label{tab:algo-comp}
    \vspace{-0.3cm}
\end{table}

As mentioned above, we provide a more efficient implementation for submodular optimization. Due to the page constraint, we defer the time complexity analysis in Appendix \ref{sec:pseudo-code}. Following the theoretical analysis, we further provide a proof-of-concept experiment to demonstrate our implementation's effectiveness advantage. Specifically, we compare our submodular optimization algorithm with two popular libraries: Submodlib \cite{kaushal2022submodlib,yang2023crest,kothawade2022prism} and DeepCore \cite{guo2022deepcore,xiao2024feature}.

For each setting, we execute each algorithm five times and report the average results. We focus on two submodular optimization strategies: the naive and stochastic greedy algorithm. The input consists of 100-dimensional vectors representing gradient information. As initial results show that Submodlib is the more effective library, we implement an alternative that replaces Submodlib’s standard similarity kernel with a custom CUDA-based kernel to verify that the performance gains primarily result from the optimization method rather than the score computation.

Table \ref{tab:algo-comp} presents the experiment results for stochastic greedy settings, while naive greedy settings is deferred to Appendix \ref{sec:naive-greedy-comp}. 
Among the evaluated methods, DeepCore performs the worst, due to its Python implementation, which introduces significant overhead. 
Submodlib demonstrates strong performance on smaller datasets (around 100 items). However, such small datasets are unsuitable for real-world applications. Our method performs less favorably on small datasets due to the high synchronization cost between CPU and GPU, which becomes a bottleneck at smaller scales. 
Nevertheless, as the dataset size increases, our method exhibits significant gains. When compared to Submodlib using a GPU-based similarity kernel, our approach achieves remarkable speedups -- up to $180.74\times$ for the naive greedy algorithm and $138.65\times$ for the stochastic greedy variant -- highlighting our efficiency in large-scale scenarios.



\section{Experiments}
\label{sec:exp}
In this section, we conduct experiments to study the effectiveness
of our method. Specifically, we are interested in answering the following
research questions (RQs):
\begin{enumerate}
    \item [\textbf{RQ1}:] How does the performance of NaCS compare to that of other established methods across various experimental settings?
    \item [\textbf{RQ2}:] What is the computational efficiency of NaCS method compared to previous methods and complete model training?
    \item [\textbf{RQ3}:] What are the effects of the key components?
    \item [\textbf{RQ4}:] To what extent do our identified coresets generalize and perform across different architectural designs?
    \item [\textbf{RQ5}:] Is NaCS compatible with LLM-based CRS models?
\end{enumerate}

\vspace{-0.3cm}
\subsection{Settings}

\subsubsection{Datasets}
We conduct our experiments on three public datasets: Criteo \cite{Criteo}, Avazu \cite{Avazu}, and KDD \cite{KDD}. In all datasets, we randomly split them into 8:1:1 as the training, validation, and test sets, respectively.
The dataset preprocessing is described in Appendix \ref{sec:dataset-preprocess}. 


\begin{table*}[thb]
    \centering
    \caption{\textbf{Comparative results with various coreset selection methods on Criteo, Avazu, and KDD under three coreset size budgets.} \textmd{``-'' denotes unachievable. The best result in each settings is marked in \textbf{bold}.}}
    \label{tab:rq1-coreset-results}
    \vspace{-0.3cm}
    \begin{adjustbox}{max width=\linewidth}
    \setlength\tabcolsep{3pt}
    \renewcommand{\arraystretch}{0.71}
    \begin{tabular}{ccc cccccc | cccccc}
    \toprule
        \multirow{2}{*}{Dataset} & \multirow{2}{*}{Backbone} & \multirow{2}{*}{Ratio} & \multicolumn{6}{c|}{AUC $\uparrow$} & \multicolumn{6}{c}{Log Loss $\downarrow$}  \\
        \cmidrule(lr){4-9} \cmidrule(lr){10-15}
        & &  & Random & SVP-CF & K-Center & CGM & Ours & Whole data 
             & Random & SVP-CF & K-Center & CGM & Ours & Whole data \\
        \midrule
        \multirow{6}{*}{Criteo} 
            & \multirow{3}{*}{DeepFM} & 0.5\% 
                        & 0.7353 & 0.7364 & 0.7162 & 0.7320 & \textbf{0.7477} & \multirow{3}{*}{0.8082} 
                        & 0.5275 & 0.5192 & 0.5410 & 0.5148 & \textbf{0.4951} & \multirow{3}{*}{0.4436} \\
                                    & & 1\% 
                        & 0.7667 & 0.7673 & 0.7620 & 0.7316 & \textbf{0.7706} & 
                        & 0.4793 & 0.4783 & 0.4859 & 0.5122 & \textbf{0.4779} \\
                                    & & 5\% 
                        & 0.7790 & 0.7792 & 0.7704 & 0.7321 & \textbf{0.7839} & 
                        & 0.4715 & 0.4719 & 0.4771 & 0.5153 & \textbf{0.4655} \\
            \cmidrule{2-15}
            & \multirow{3}{*}{DCNv2}  & 0.5\% 
                        & 0.7211 & 0.7227 & 0.6983 & 0.7288 & \textbf{0.7469} & \multirow{3}{*}{0.8117} 
                        & 0.5514 & 0.5532 & 0.6026 & 0.5160 & \textbf{0.5061} & \multirow{3}{*}{0.4402} \\
                                    & & 1\% 
                        & 0.7673 & 0.7685 & 0.7665 & 0.7261 & \textbf{0.7707} & 
                        & 0.4779 & 0.4776 & 0.4789 & 0.5317 & \textbf{0.4758} \\
                                    & & 5\% 
                        & 0.7853 & 0.7868 & 0.7730 & 0.7234 & \textbf{0.7910} &
                        & 0.4634 & 0.4626 & 0.4742 & 0.5301 & \textbf{0.4611} \\
        \midrule
        \multirow{6}{*}{Avazu} 
            & \multirow{3}{*}{DeepFM} & 0.5\% 
                        & 0.6727 & 0.6894 & 0.6617 & - & \textbf{0.7071} & \multirow{3}{*}{0.7852} 
                        & 0.4961 & 0.5733 & 0.5204 & - & \textbf{0.4732} & \multirow{3}{*}{0.3771} \\
                                    & & 1\% 
                        & 0.7145 & 0.7117 & 0.6776 & 0.6980 & \textbf{0.7383} & 
                        & 0.4205 & 0.4179 & 0.4766 & 0.4946 & \textbf{0.4049} \\
                                    & & 5\% 
                        & 0.7419 & 0.7483 & 0.7197 & 0.7015 & \textbf{0.7502} &
                        & 0.4027 & 0.3973 & 0.4122 & 0.6169 & \textbf{0.3962} \\
            \cmidrule{2-15}
            & \multirow{3}{*}{DCNv2}  & 0.5\% 
                        & 0.6694 & 0.6781 & 0.6714 & - & \textbf{0.6905} & \multirow{3}{*}{0.7891} 
                        & 0.6210 & 0.6138 & 0.6239 & - & \textbf{0.4414} & \multirow{3}{*}{0.3750} \\
                                    & & 1\% 
                        & 0.6953 & 0.7103 & 0.6745 & 0.6872 & \textbf{0.7365} & 
                        & 0.4459 & 0.4359 & 0.5044 & 0.4385 & \textbf{0.4036} \\
                                    & & 5\% 
                        & 0.7427 & 0.7410 & 0.7145 & 0.6890 & \textbf{0.7522} & 
                        & 0.4003 & 0.4012 & 0.4158 & 0.4798 & \textbf{0.3955} \\  
        \midrule
        \multirow{6}{*}{KDD} 
            & \multirow{3}{*}{DeepFM} & 0.5\% 
                        & 0.7129 & 0.7127 & 0.7068 & - & \textbf{0.7298} & \multirow{3}{*}{0.7976} 
                        & 0.1677 & 0.1682 & 0.1718 & - & \textbf{0.1651} & \multirow{3}{*}{0.1527} \\
                                    & & 1\% 
                        & 0.7242 & 0.7238 & 0.7108 & 0.7042 & \textbf{0.7443} & 
                        & 0.1658 & 0.1659 & 0.1692 & 0.2506 & \textbf{0.1633} \\
                                    & & 5\% 
                        & 0.7478 & 0.7445 & 0.7364 & 0.7053 & \textbf{0.7729} &
                        & 0.1621 & 0.1626 & 0.1654 & 0.2111 & \textbf{0.1582} \\
            \cmidrule{2-15}
            & \multirow{3}{*}{DCNv2}  & 0.5\% 
                        & 0.7120 & 0.7107 & 0.6944 & - & \textbf{0.7300} & \multirow{3}{*}{0.7995} 
                        & 0.1679 & 0.1685 & 0.1745 & - & \textbf{0.1650} & \multirow{3}{*}{0.1523} \\
                                    & & 1\% 
                        & 0.7242 & 0.7141 &  0.7082 & 0.6911 & \textbf{0.7441} & 
                        & 0.1657 & 0.1675 &  0.1690 & 0.3417 & \textbf{0.1630} \\
                                    & & 5\% 
                        & 0.7470 & 0.7452 & 0.7346 & 0.7016 & \textbf{0.7737} & 
                        & 0.1624 & 0.1624 & 0.1660 & 0.2650 & \textbf{0.1578} \\ 
    \bottomrule
    \end{tabular}
    \end{adjustbox}

\end{table*}

\vspace{-0.15cm}
\subsubsection{Metrics} 
We evaluate all methods using two commonly used metrics in CTR prediction community \cite{barsctr2021}: AUC (Area under the ROC curve) and Log Loss. In CTR prediction, a difference of 0.1\% in AUC is generally considered significant \cite{barsctr2021}. The lower LogLoss indicates better performance, while a higher AUC suggests more accurate recommendations.  

\vspace{-0.15cm}
\subsubsection{Implementation Details} 
We compare NaCS against other data reduction methods, namely: Random, SVP-CF \cite{svpcf2022}, K-Center \cite{sener2017active}, and CGM \cite{wang2023gradient}, whose details are specified in Appendix \ref{sec:baselines}. 
All methods are tested with two backbones: 
DeepFM \cite{deepfm2017} and DCNv2 \cite{dcnv2}, whose hyperparameters are specified in Appendix \ref{sec:hyperparameters}. 
Following prior studies \cite{barsctr2021}, we employ early stopping to prevent overfitting and mitigate error accumulation in our progressive self-correction approach (Appendix \ref{sec:learn-curve}).

\begin{figure*}[t!]
    \centering
    \ref{legend_v2}
    \vspace{0.08cm}
    
    \subcaptionbox{Criteo}{


\begin{tikzpicture}
\tikzstyle{every node}=[font=\scriptsize]
\pgfplotsset{set layers}
    \begin{axis}[
        width=0.37\linewidth,
        height=4.3cm,
        ybar,
        ylabel={Time (s)},
        enlargelimits=0.15,
        legend style={at={(0.5,-0.22)},
            anchor=north,legend columns=-1, font=\scriptsize},
        bar width=8pt,
        xmin={[normalized]-0.2},
        xmax={[normalized]2},
        xlabel={Subset size},
        scaled ticks=base 10:-3,
        symbolic x coords={0.5\%, 1\%, 5\%},
        xtick=data,
        legend to name=legend_v2,
    ]
    
        \addplot+[nodes near coords] coordinates {
            (0.5\%, 851)
            (1\%, 1299)
            (5\%, 1380)
        };
        \addplot+[nodes near coords] coordinates {
            (0.5\%, 499)
            (1\%, 499)
            (5\%, 499)
        };
        \addplot+[color=green!60!black,stack plots=y,bar shift auto=2pt] coordinates {
            (0.5\%, 158)
            (1\%, 165)
            (5\%, 370)
        };
        \addplot+[color=green!60!black,pattern=crosshatch,pattern color=green!60!black, stack plots=y,bar shift auto=-4.6pt, nodes near coords] coordinates {
            (0.5\%, 24)
            (1\%, 31)
            (5\%, 108)         
        };


    \path (axis cs:{[normalized]1},736) coordinate (aux);
    
    \addlegendimage{black, line legend, dashed}
    \legend{CGM, SVP, Select, Denoise, Original}
  \end{axis}
  
  \draw[dashed, black] (current axis.west|-aux) -- (current axis.east|-aux);



\end{tikzpicture}}
    \hfill
    \subcaptionbox{Avazu}{


\begin{tikzpicture}
\tikzstyle{every node}=[font=\scriptsize]
\pgfplotsset{set layers}
    \begin{axis}[
        width=0.37\linewidth,
        height=4.3cm,
        ybar,
        enlargelimits=0.15,
        legend style={at={(0.5,-0.22)},
            anchor=north,legend columns=-1},
        bar width=8pt,
        xmin={[normalized]-0.2},
        xmax={[normalized]2},
        xlabel={Subset size},
        scaled ticks=base 10:-3,
        symbolic x coords={0.5\%, 1\%, 5\%},
        xtick={0.5\%, 1\%, 5\%},
    ]
    
        \addplot+[nodes near coords] coordinates {
            (1\%, 983)
            (5\%, 998)
        };
        \addplot+[nodes near coords] coordinates {
            (0.5\%, 570)
            (1\%, 570)
            (5\%, 570)
        };
        \addplot+[color=green!60!black,stack plots=y,bar shift auto=2pt] coordinates {
            (0.5\%, 169)
            (1\%, 191)
            (5\%, 340)
        };        
        \addplot+[color=green!60!black,pattern=crosshatch,pattern color=green!60!black, stack plots=y,bar shift auto=-4.6pt, nodes near coords] coordinates {
            (0.5\%, 53)
            (1\%, 57)
            (5\%, 135)         
        };

    
    \path (axis cs:{[normalized]1},694) coordinate (aux);
    \addlegendimage{black, line legend, dashed}
    
  \end{axis}
  
  \draw[dashed, black] (current axis.west|-aux) -- (current axis.east|-aux);



\end{tikzpicture}}
    \hfill
    \subcaptionbox{KDD}{\begin{tikzpicture}
\tikzstyle{every node}=[font=\scriptsize]
\pgfplotsset{set layers}
    \begin{axis}[
        width=0.37\linewidth,
        height=4.3cm,
        ybar,
        ymin=1200, ymax=12000,
        enlargelimits=0.15,
        legend style={at={(0.5,-0.22)},
            anchor=north,legend columns=-1},
        bar width=8pt,
        xmin={[normalized]-0.2},
        xmax={[normalized]2},
        xlabel={Subset size},
        scaled ticks=base 10:-3,
        symbolic x coords={0.5\%, 1\%, 5\%},
        xtick={0.5\%, 1\%, 5\%},
    ]
    
        \addplot+[nodes near coords] coordinates {
            (1\%, 9693)
            (5\%, 10852)
        };
        \addplot+[nodes near coords,
        every node near coord/.append style={xshift=1pt}
        ] coordinates {
            (0.5\%, 2609)
            (1\%, 2609)
            (5\%, 2609)
        };
        \addplot+[color=green!60!black,stack plots=y,bar shift auto=2pt] coordinates {
            (0.5\%, 278)
            (1\%, 297)
            (5\%, 546)
        };
        \addplot+[color=green!60!black,pattern=crosshatch,pattern color=green!60!black, stack plots=y,bar shift auto=-4.6pt, nodes near coords] coordinates {
            (0.5\%, 99)
            (1\%, 115)
            (5\%, 208)         
        };


    \path (axis cs:{[normalized]1},10201) coordinate (aux);
    
    \addlegendimage{black, line legend, dashed}
  \end{axis}
  
  \draw[dashed, black] (current axis.west|-aux) -- (current axis.east|-aux);

\end{tikzpicture}}
    \hfill
    \vspace{-0.3cm}
    \caption{\textbf{Runtime relative to the full dataset of DCNv2 backbone with various budgets.} \textmd{``Select'' and ``Denoise'' represent the runtime of our two steps. ``Original'' represents the training time on the original dataset.} 
    }
    \label{fig:rq2}
    \vspace{-0.2cm}
\end{figure*}

\vspace{-0.3cm}
\subsection{Overall Performance (RQ1)}

Table \ref{tab:rq1-coreset-results} shows the performance of NaCS compared with other baselines.
Across all settings, our method significantly outperforms all other baselines, with a more prominent gap when data availability is limited. 
As with a smaller coreset size, we require a better selection strategy to achieve a good result. 
NaCS retains approximately 95\% AUC on Criteo and 93\% on Avazu and KDD using only 1\% of the dataset.
These results highlight our strength in selecting informative samples on a compact budget.

Interestingly, while commonly considered the weaker backbone, DeepFM generally outperforms DCNv2 when trained on smaller coresets.
One plausible explanation is that simpler architectures, such as DeepFM, are less prone to overfitting,  generalize better, and lead to more stable coreset selections. 
This insight emphasizes the necessity of aligning model complexity with the data budget.

Moreover, random is still a competitive baseline compared to other proposed approaches, which is a similar observation to recent studies \cite{qin2025d2k,lin2024data,wang2023gradient}.
Additionally, although CGM can use very few samples, each sample encodes a dense representation, which limits its ability to scale under strict data budgets.

\subsection{Efficiency Analysis (RQ2)}

For method efficiency, we conduct the experiments to benchmark the time required to generate and to train on the result coreset. For this experiment, we employ a GPU workstation with an i7-13700K CPU, NVIDIA RTX A5000 GPU, and 32GB RAM.

As shown in Figure \ref{fig:rq2} and Appendix \ref{sec:exp-extra}, our method provides substantial gains in training efficiency. 
In 3-4 minutes, we can select 1\% of the data for the Criteo and Avazu datasets, while training on the full datasets requires over 12.5 minutes and 11 minutes. 
Furthermore, NaCS consistently outperforms previous approaches in efficiency. This advantage stems from avoiding iterative backpropagation across the entire dataset, unlike these methods, which require either constant-time score assignment (e.g., SVP-CF) over all samples or expensive bi-level optimization schemes (e.g., CGM).

In NaCS, the coreset selection step accounts for a considerable runtime, as it requires forward passes over the entire dataset to compute last-layer gradients, submodular optimization, and progressive training on the coreset.
In contrast, the denoising step only requires forward passes on a negligible subset of the data.

\subsection{Ablation Study (RQ3)}

In the below part, we verify the effectiveness of our three main components. For this experiment, we employ 1\% coreset size with DCNv2 and DeepFM backbones as 1\% subset provides a good trade-off between efficiency and performance. 

\subsubsection{Coreset Objective} In this experiment, we employ two additional baselines: Random and set cover \cite{cormode2010set}, which aim to maximize the number of features in the coreset. For a fair comparison, we denoise the resulting coresets of these methods similarly to our proposed approach. As shown in Table \ref{tab:rq3-merge}a, our proposed approach performs the best on all three datasets. Moreover, the random approach also benefits from the denoising process.



\begin{table*}[tbh]
    \centering
    \caption{Ablation study results. \textmd{In each settings, the best result is highlighted in \textbf{bold} and the second best is \underline{underlined}.} }
    \vspace{-0.3cm}
    \label{tab:rq3-merge}
    \renewcommand{\arraystretch}{0.8}
    \setlength\tabcolsep{2.5pt}
    \begin{tabular}{ll cccc cccc cccc}
    \toprule
        \multirow{3}{*}{Section} & \multirow{3}{*}{Method} & \multicolumn{4}{c}{Criteo} & \multicolumn{4}{c}{Avazu} & \multicolumn{4}{c}{KDD} \\
        \cmidrule(lr){3-6} \cmidrule(lr){7-10} \cmidrule(lr){11-14}
        & & \multicolumn{2}{c}{DeepFM} & \multicolumn{2}{c}{DCNv2} & \multicolumn{2}{c}{DeepFM} & \multicolumn{2}{c}{DCNv2} & \multicolumn{2}{c}{DeepFM} & \multicolumn{2}{c}{DCNv2} \\
        \cmidrule(lr){3-4} \cmidrule(lr){5-6} \cmidrule(lr){7-8} \cmidrule(lr){9-10} \cmidrule(lr){11-12} \cmidrule(lr){13-14}
        & & AUC & Log Loss & AUC & Log Loss 
        & AUC & Log Loss & AUC & Log Loss 
        & AUC & Log Loss & AUC & Log Loss \\ 
        \midrule
        (a) Coreset & Set Cover & 0.7460 & 0.4928 & 0.7520 & 0.4935 & 0.7067 & 0.4168 & 0.7098 & 0.4192 & 0.7229 & 0.1674 & 0.7262 & 0.1671  \\
        Objective & Random & 0.7693 & \textbf{0.4757} & 0.7696 & \underline{0.4765} & 0.7186 & 0.4141 & 0.6948 & 0.4279 & 0.7281 & 0.1654 & 0.7265 & 0.1659 \\
        
        \midrule
        & BCE & 0.7611 & 0.4834 & 0.7625 & 0.4811 & 0.7254 & 0.4096 & 0.7246 & 0.4095 & 0.7391 & \underline{0.1632} & 0.7408 & \underline{0.1630} \\
        (b) Loss & T-CE & 0.7451 & 0.5597 & 0.7615 & 0.6123 & 0.7185 & 0.5279 & 0.7295 & 0.4812 & 0.7311 & 0.1702 & 0.7320 & 0.1685 \\
        & R-CE & 0.7672 & 0.4812 & 0.7681 & 0.4786 & 0.7298 & 0.4104 & 0.7322 & 0.4076 & 0.7440 & 0.1704 & 0.7435 & 0.1724 \\
        
        \midrule 
        (c) Denoise & w/o denoise & 0.7675 & 0.4782 & 0.7659 & 0.4785 & 0.7346 & 0.4052 & 0.7308 & 0.4068 & 0.7390 & 0.1634 & 0.7422 & \underline{0.1630} \\
        approaches & Full model & \underline{0.7695} & \underline{0.4765} & \underline{0.7705} & 0.4770 & \textbf{0.7395} & \textbf{0.4025} & \textbf{0.7379} & \textbf{0.4032} & \textbf{0.7506} & \textbf{0.1620} & \textbf{0.7482} & \textbf{0.1628} \\
        \midrule
        & NaCS & \textbf{0.7706} & 0.4779 & \textbf{0.7707} & \textbf{0.4758} & \underline{0.7383} & \underline{0.4049} & \underline{0.7365} & \underline{0.4036} & \underline{0.7443} & 0.1633 & \underline{0.7441} & \underline{0.1630} \\
    \bottomrule
    \end{tabular}

\end{table*}

\subsubsection{Loss Function} Table \ref{tab:rq3-merge}b presents ablation studies for BCE and other loss-based denoising approaches \cite{wang2021wsdm_denoise} compared to our proposed approach. The results show that both our method and R-CE improve the performance over the traditional log loss approach, with our method achieving the best performance. On the other hand, T-CE performs poorly in our settings. One possible hypothesis is that they discard positive interactions too severely, leading to poor recommendations in small datasets.



\subsubsection{Denoising Effectiveness} Table \ref{tab:rq3-merge}c depicts the ablation study on data denoising. In ``Full model'', instead of employing the model trained on the coreset as a denoiser, we utilize the model trained on the whole dataset. This ``Full model'' can be accessed if we have more computing resources to produce a better coreset. The experiment results show that denoising the dataset benefits the model on all datasets. Moreover, we can employ the ``Full model'' to further enhance the coreset denoising process.



\subsection{Cross Architecture Performance (RQ4)}

Table \ref{tab:rq4} presents the AUC performance of various architectures when evaluated with a 1\% and 5\% dataset budget on three datasets. 
For this comparison, we further include FinalMLP \cite{mao2023finalmlp} as another backbone in model training.
The performance gap between backbones is minor, highlighting our coreset's transferability. 
At a 1\% dataset size, the DeepFM backbone yields a better coreset, while DCNv2 performs better at a 5\% size. 
This may be because DeepFM, as a smaller model, is less prone to overfitting with limited samples, whereas DCNv2's superior representation power becomes advantageous with a larger budget. 
Generally, stronger backbones outperform weaker ones on the same datasets. Except for the 1\% data budget scenario on KDD, FinalMLP consistently outperforms the other baselines. 

\begin{table}[t!]
    \centering
    \caption{Cross-architecture AUC performance with a 1\% and 5\% dataset budget on the Criteo, Avazu, and KDD datasets. \textmd{``Backbones'' refer to downstream models.}}
    \label{tab:rq4}
    \vspace{-.3cm}
    \begin{adjustbox}{max width=\linewidth}
    \setlength\tabcolsep{2pt}
    \renewcommand{\arraystretch}{0.8}
    \begin{tabular}{cl cc cc cc}
    \toprule
        Data & \multirow{2}{*}{Backbones} & \multicolumn{2}{c}{Criteo} & \multicolumn{2}{c}{Avazu} & \multicolumn{2}{c}{KDD} \\
        \cmidrule(lr){3-4} \cmidrule(lr){5-6} \cmidrule(lr){7-8} 
        size & & DeepFM & DCNv2 & DeepFM & DCNv2 & DeepFM & DCNv2\\
    \midrule
       \multirow{3}{*}{1\%} & DeepFM & 0.7706 & 0.7700 & 0.7383 & 0.7378 & 0.7443 & 0.7435\\
        & DCNv2 & 0.7696 & 0.7707 & 0.7396 & 0.7365 & 0.7410 & 0.7441 \\
        & FinalMLP & 0.7717 & 0.7708 & 0.7398 & 0.7388 & 0.7427 & 0.7417  \\
     \midrule
       \multirow{3}{*}{5\%} & DeepFM & 0.7839 & 0.7839 & 0.7502 & 0.7519 & 0.7729 & 0.7735 \\
        & DCNv2 & 0.7903 & 0.7910 & 0.7493 & 0.7522 & 0.7725 & 0.7737\\
        & FinalMLP & 0.7910 & 0.7912 & 0.7554 & 0.7547 & 0.7736 & 0.7741 \\
    \bottomrule
    \end{tabular}
    \end{adjustbox}

\end{table}

\subsection{Compatibility with LLMs (RQ5)}
Since our method relies solely on model gradients, it can be easily extended to other recommendation settings. 
Given the growing interest in LLM-based models, we apply our approach to a text-based recommendation scenario. 
To this end, we implement our algorithm on the Legommenders  \cite{liu2025legommenders} and evaluate our performance on the MIND dataset. As is standard practice with this dataset, we perform negative sampling, generating four negative samples for each positive instance. 
Table \ref{tab:rq7} shows the experiment results.

\begin{table}[h]
    \vspace{-.1cm}
    \centering
    \caption{Performance of coreset selection methods on MIND dataset. ``N'' is short for NDCG.}
    \vspace{-.3cm}
    \begin{adjustbox}{max width=\linewidth}
    \setlength\tabcolsep{3pt}
    \renewcommand{\arraystretch}{0.7}
    \begin{tabular}{cclcccc}
        \toprule
        Backbone & Data size & Method & GAUC & MRR & N@5 & N@10  \\
        \midrule
        \multirow{5}{*}{NAML} & \multirow{1}{*}{100\%} & Whole data & 0.6704 & 0.2931 & 0.3263 & 0.3901 \\
        \cmidrule(lr){2-7}
        & \multirow{2}{*}{1\%} & Random & 0.5883 & 0.2454 & 0.2658 & 0.3290 \\
        & & NaCS & 0.5984 & 0.2497 & 0.2740 & 0.3364  \\
        \cmidrule(lr){2-7}
        & \multirow{2}{*}{5\%} & Random & 0.6373 & 0.2733 & 0.3006 & 0.3650 \\
        & & NaCS & 0.6422 & 0.2744 & 0.3060 & 0.3677 \\
        \midrule
        \multirow{5}{*}{ONCE} & 100\% & Whole data & 0.6874 & 0.3106 & 0.3522 & 0.4131 \\
        \cmidrule(lr){2-7}
        & \multirow{2}{*}{1\%} & Random & 0.6004 & 0.2512 & 0.2720 & 0.3357 \\
        & & NaCS & 0.6273 & 0.2665 & 0.2955 & 0.3562 \\
        \cmidrule(lr){2-7}
        & \multirow{2}{*}{5\%} & Random & 0.6552 & 0.2879 & 0.3246 & 0.3851 \\
        & & NaCS & 0.6586 & 0.2935 & 0.3321 & 0.3921 \\
        \bottomrule
    \end{tabular}
    \end{adjustbox}
    \label{tab:rq7}
\end{table}

In text-based recommendation, our method demonstrates consistent improvement over random selection, despite a smaller amount compared to the traditional recommendation task. One possible explanation is that the influence of pretrain text embedding.
Interestingly, despite MIND being a relatively small dataset, the larger model, ONCE, outperforms the smaller model, NAML, across both low and high coreset budget settings, instead of overfitting. This contrasts with trends observed in traditional category-based recommendation tasks. A possible explanation is the stronger generalization capability of pretrained text-based models, particularly large language models (LLMs), which may reduce overfitting on smaller datasets.

\section{Conclusion}
\label{sec:conclusion}
In this paper, we propose a noise-aware coreset selection framework for content-based recommendation systems. Our proposed approach first selects a small subset of samples to mimic the full dataset gradient by applying submodular optimization. Concurrently, we assign new labels for each sample with a progressive self-correction approach. Then, through Monte Carlo Dropout, we efficiently identify uncertain samples from the coreset, resulting in a clean and performant coreset. Extensive experiments show that our method is not only more efficient but also more accurate than previous state-of-the-art approaches.

\begin{acks}
The Australian Research Council partially supports this work under Grant No. FT210100624, DE230101033, DP240101108, DP260100326, LP230200892 and LP240200546.
\end{acks}

\bibliographystyle{ACM-Reference-Format}
\balance
\bibliography{00_main}

\appendix
\section{Extended Related Work}
\label{sec:extended-rw}

\subsection{Recommendation Efficiency}
Various methods were proposed to improve the recommendation model efficiency, including: storage efficiency \cite{shaver2025,cerp2023,legcf2024,yunke_continous2023,bet2024}, inference efficiency \cite{zhang2025hmamba,xia2022device}, and training efficiency \cite{xia2023efficient,gao2025graph,qu2024sparser,qu2025efficient}. Our work focuses on training efficiency. 
APG \cite{yan2022apg} dynamically generates parameters based on individual input samples to enhance model adaptability. 
BAGPIPE \cite{agarwal2023bagpipe} discovers that the embedding queries and updates are the main bottleneck for training recommendation models. 
Our proposed framework is orthogonal to these approaches, as it focuses on data efficiency and is not affected by the model structure or how the model weight is updated.

Various data-centric strategies were also proposed.
IntellectReq \cite{lv2024intelligent} reduces communication costs for on-device recommendation \cite{yin2024device} by using a Mis-Recommendation Detector and a variational inference-based Distribution Mapper to predict whether cloud-based model updates will significantly improve performance.
Distill-CF \cite{sachdeva2022infinite} proposes a dataset distillation for collaborative filtering using bi-level optimization. \camcomment{To reduce the computational cost for the inner loop, they employ a custom backbone that has a closed-form solution.}
\camcomment{Recently, various data-efficient methods for text-based recommendation were also proposed \cite{wu2025leveraging,lin2024data}.
However, these methods focus on ranking-based recommendation tasks and cannot be directly applied to content-based recommendation.}

\subsection{Coreset Selection}

Coreset selection is a well-studied topic in both traditional
machine learning and deep learning. 
Various heuristics have also been explored for selecting the coresets. They either attempt to train a full model \cite{birodkar2019semantic} or a smaller proxy model \cite{coleman2019selection}. Based on this training process, they assign heuristic scores, such as the gradient norm \cite{paul2021deep}, the highest uncertainty \cite{coleman2019selection},  or the number of forgetting events \cite{tonevaempirical2019}. However, these methods  require training on the whole dataset and cannot perform a trade-off between performance and efficiency, as they commonly require assigning scores for every sample.

Coreset selection can be formulated as a bilevel optimization problem, where the outer objective focuses on selecting a subset of data (samples or weights), and the inner objective involves optimizing model parameters on this subset \cite{guo2022deepcore}. 
Several methods exemplify this framework, including cardinality-constrained bilevel optimization for continual learning \cite{borsos2020coresets}, Retrieve \cite{killamsetty2021retrieve} for semi-supervised learning\camcomment{, and Glister \cite{killamsetty2021glister} for supervised and active learning}. 
However, these bi-level optimization approaches are costly and are not suitable for large scale dataset.

Our method follows the submodular optimization approach.
CRAIG \cite{mirzasoleiman2020coresets} is one of the first coreset selection methods based on submodular optimization, providing a general framework that our work builds upon. CREST \cite{yang2023crest} extends this by focusing on when to select the coreset, rather than using a fixed selection step like CRAIG. CoLM \cite{dang2025colm} further explores submodular optimization in the context of LLM-based models.
Compared to previous methods, our proposed method involves two novel key components: progressive label correction and denoise module, making it more suitable for noisy recommendation datasets. Moreover, we include a novel implementation for stochastic greedy, resulting in more than $100\times$ speed up than previous implementation.

\subsection{Implicit Feedback Denoising}


It is well-known that implicit feedback is noisy.
Recent studies \cite{wang2021wsdm_denoise} have highlighted the substantial noise in implicit feedback datasets due to their characteristics.
To tackle this problem, numerous approaches have been proposed to denoise implicit feedback. 
Initial work leverages auxiliary information, namely user behaviors (e.g. ``dwell time'' \cite{kim2014modeling}, ``like'' \cite{bian2021denoising}) and item side-information.
However, these side features can be costly and challenging to collect, restricting their real-world usage, and creating a large demand for identifying noisy samples without external signals. 

Wang et al. \cite{wang2021wsdm_denoise} discovered that the noisy samples can be found through the loss values, and we can efficiently reduce noise by modifying the loss function.
\camcomment{BOD \cite{wang2023efficient} employs bi-level optimization, learning the sample weight and the model weight at the same time.}
DCF \cite{he2024dcf} proposes both data pruning and label correction by observing the loss value while training the model. 
\camcomment{SVV \cite{zhang2025shapley} applies Shapley values to quantify the contribution of each sample to the overall loss value, effectively identifying low-contribution samples.}
Although numerous solutions have been proposed to denoise implicit feedback, we are the first to integrate it into the coreset selection framework. While the denoise-focused approaches attempt to remove a small portion of the dataset, we prune most samples from the dataset.

\section{Pseudo Code}
\label{sec:pseudo-code}

\RestyleAlgo{ruled}
\LinesNumbered
\begin{algorithm}[t]
\setstretch{0.7}
\caption{Torch-based Batch Submodular Optimization}
\label{alg:submod}
\SetKwRepeat{Do}{do}{while}
\SetKwComment{Comment}{/* }{ */}

\KwData{Batches $\{ \mathcal{D}_1, \mathcal{D}_2, \dots, \mathcal{D}_B \}$, the subset size $k$, Batch size $N = \lvert \mathcal{D}_i \rvert = \lvert\mathcal{D}_j\rvert, \forall i, j \in \left[1, B \right]$}
\KwResult{Subsets $\{ \mathcal{S}_1, \mathcal{S}_2, \mathcal{S}_3, \dots, \mathcal{S}_B \}$}

$\forall i \in [1, B], \mathbf{S}_i \gets $ Calculate similarity kernel of $\mathcal{D}_i$\; 
$\forall i \in [1, B], \mathcal{S}_i \gets \emptyset$\;

$s \gets \lceil N \times \log(1 / \epsilon) / k   \rceil$\;

$\mathbf{W} \gets 1^{B \times N}$   \;

Values $\mathbf{V} \gets \mathbb{-\inf}^{B \times N}$\;
Value Sum $\mathbf{u} \gets 0^{B}$\;

\For{$j=1, 2, \dots, k$}{
    \For{$i=1, 2, \dots, B$}{
        $\mathcal{T}_i \sim$ Multinomial($w$, size=$s$)\;
        $\mathbf{G}_i \gets \max(\mathbf{S}_i\left[ \mathcal{T}_i \right], \mathbf{V}_i) - \mathbf{u}_i$\;
        $\text{Item}_i \gets \arg\max\mathbf{G}_i$\;
        $\mathcal{S}_i \gets \mathcal{S}_i \cup \{ \text{Item}_i \}$\;
        
        $\mathbf{V}_i \gets \max(\mathbf{V}_i, \mathbf{S}_{\text{Item}_i})$, $\mathbf{u}_i \gets \sum \mathbf{V}_i$\;
        $\mathbf{W}_{i, \text{Item}_i} \gets 0$\;
    }

}

\end{algorithm}

Our whole framework is summarized in Algorithm \ref{alg:main}. Algorithm \ref{alg:submod} shows our main submodular algorithm, which corresponding to line \textbf{7} in Algorithm \ref{alg:main}.
In the following part, we provide the time complexity analysis for our submodular optimization approaches. First, we compute the similarity between each pair of items in batches (line \textbf{1}), resulting in an $O(BN^2)$ time complexity. Then, we initialize various required variables (Lines \textbf{2-6}).  In the stochastic greedy algorithm, at each iteration, we sample a small random subset with the size of $s$ (line \textbf{9}), and select the element with the largest marginal gain from this subset (lines \textbf{10-12}) rather than inspecting all candidates. Due to submodularity, this ensures that with high probability, the sampled subset contains a near-best element, enabling an expected $1 - 1/e - \varepsilon$ approximation guarantee with only a time complexity of $O\left( \vert \mathcal{D} \vert \log \frac{1}{\varepsilon} \right)$, independent from the cardinality constraint $k$. It is worth noting that, in actual implementation, line \textbf{1} and lines \textbf{8-15} will be computed in parallel, thus significantly boosting the efficiency.

\section{Experiment Settings}


\subsection{Dataset}
\label{sec:dataset-preprocess}

For Avazu and Criteo dataset, we directly employ the preprocessed dataset from \cite{tran2024thorough}.
Following previous works \cite{shaver2025,optembed2022}, we utilize OOV tokens to represent infrequent features (appear less than 10 times) for KDD dataset.
Table \ref{tab:data-stats} shows core statistics of pre-processed datasets.

\begin{table}[tbh]
    \centering
    \vspace{-0.2cm}
    \caption{Statistics of the preprocessed datasets.}
    \vspace{-0.3cm}
    \renewcommand{\arraystretch}{0.8}
    \begin{tabular}{lrrr}
    \toprule
    Name & \#Interactions & \#Features & \#Fields \\
    \midrule
    Criteo & 45,840,617 & 1,086,810 & 39 \\
    Avazu & 40,428,967 & 4,428,511 & 22 \\
    KDD & 149,639,105 & 6,019,761 & 11 \\
    \bottomrule
    \end{tabular}
    \label{tab:data-stats}
    \vspace{-0.3cm}
\end{table}

\subsection{Baselines}
\label{sec:baselines}
We compared our proposed approach with the following baselines:
\begin{itemize}
    \item \textbf{Random}: We select random samples from the original data.
    \item \textbf{SVP-CF} \cite{svpcf2022}: We train a Factorization Machine model as the proxy model to calculate the importance score (inverse AUC) through the training process. Then we choose the samples with the highest inverse AUC scores.
    \item \textbf{K-Center} \cite{sener2017active}: The K-Center algorithm selects a coreset by minimizing the maximum distance between any dataset point and its nearest coreset sample. As it is a submodular optimization problem, we apply the stochastic greedy algorithm to smaller dataset subsets for efficiency. 
    \item \textbf{CGM} \cite{wang2023gradient}: Following the proposed method, we train a synthetic dataset to mimic the original dataset's gradient. For a fair comparison, we calculate the storage required by coresets and set a similar budget for condensed datasets. 
\end{itemize}

\begin{figure*}[thb]
    \centering

    \ref{legend_v1}
    \vspace{0.1cm}
    
    \subcaptionbox{Criteo}{\begin{tikzpicture}
    \pgfplotsset{set layers}
    \begin{axis}[
        ybar stacked,
        bar width=15pt,
        axis y line*=left,
        width=0.31\linewidth,
        height=4.1cm,
        ymin=0, ymax=1200,
        xlabel={Subset size},
        ylabel={Time (s)},
        legend style={at={(0.5,-0.22)},
            anchor=north,legend columns=-1, font=\scriptsize},
        legend to name=legend_v1,
        legend entries={Select, Denoise, Train, Performance},
        scaled ticks=base 10:-3,
        symbolic x coords={0.5\%, 1\%, 5\%, Full},
        xtick=data,
        ymajorgrids=true,
    ]

    \addplot coordinates {(0.5\%, 158) (5\%, 370) (1\%, 165) (Full, 0)};
    \addplot coordinates {(0.5\%, 24) (5\%, 108) (1\%, 31) (Full, 0)};
    \addplot coordinates {(0.5\%, 137) (5\%, 188) (1\%, 138) (Full, 736)};

    \addlegendimage{red, line legend, solid, mark=square*}

    \node[above] at (axis cs:1\%,334) {{\small 5.6m}};
    \node[above] at (axis cs:5\%,666) {{\small 11.1m}};
    \node[above] at (axis cs:0.5\%,319) {{\small 5.3m}};
    \node[above] at (axis cs:Full,736) {{\small 12.3m}};
    \end{axis}

    \begin{axis}[
        axis y line*=right,
        axis x line=none, 
        ymin=0, ymax=120,
        yticklabel style={align=right},
        width=0.31\linewidth,
        height=4.1cm,
        symbolic x coords={0.5\%, 1\%, 5\%, Full},
    ]

    \addplot[color=red, mark=square*] coordinates {(0.5\%, 92) (1\%, 95) (5\%, 97) (Full, 100)};
    \node[below] at (axis cs:1\%,95) {95\%};
    \node[below] at (axis cs:5\%,97) {97\%};
    \node[below] at (axis cs:0.5\%,92) {92\%};
    \end{axis}
    
\end{tikzpicture}}
    \subcaptionbox{Avazu}{\begin{tikzpicture}
    \begin{axis}[
        ybar stacked,
        axis y line*=left,
        bar width=15pt,
        width=0.31\linewidth,
        height=4.1cm,
        ymin=0, ymax=1000,
        xlabel={Subset size},
        legend style={at={(0.5,-0.25)},
            anchor=north,legend columns=-1, font=\scriptsize},
        scaled ticks=base 10:-3,
        symbolic x coords={0.5\%, 1\%, 5\%, Full},
        xtick=data,
        ymajorgrids=true,
    ]

    \addplot coordinates {(0.5\%, 169) (5\%, 340) (1\%, 191) (Full, 0)}; \label{pgfplots:Select1}
    \addplot coordinates {(0.5\%, 53) (5\%, 135) (1\%, 57) (Full, 0)}; \label{pgfplots:Denoise1}
    \addplot coordinates {(0.5\%, 146) (5\%, 137) (1\%, 192) (Full, 694)};

    \addlegendimage{red, line legend, solid, mark=square*}

    \node[above] at (axis cs:0.5\%,368) {{\small 6.1m}};
    \node[above] at (axis cs:1\%,440) {{\small 7.3m}};
    \node[above] at (axis cs:5\%,612) {{\small 10.2m}};
    \node[above] at (axis cs:Full,694) {{\small 11.6m}};
    \end{axis}

    \begin{axis}[
        axis y line*=right,
        axis x line=none, 
        ymin=0, ymax=120,
        yticklabel style={align=right},
        width=0.31\linewidth,
        height=4.1cm,
        symbolic x coords={0.5\%, 1\%, 5\%, Full},
    ]

    \addplot[color=red, mark=square*] coordinates {(0.5\%, 87.5) (1\%, 93.47) (5\%, 95.3) (Full, 100)};
    \node[above] at (axis cs:0.5\%,87.5) {88\%};
    \node[above] at (axis cs:1\%,93.47) {93\%};
    \node[above] at (axis cs:5\%,95.3) {95\%};
    \end{axis}
    
\end{tikzpicture}}
    \subcaptionbox{KDD}{\begin{tikzpicture}
    \pgfplotsset{set layers}
    \begin{axis}[
        ybar stacked,
        bar width=15pt,
        axis y line*=left,
        width=0.32\linewidth,
        height=4.1cm,
        ymin=0, ymax=12000,
        xlabel={Subset size},
        legend style={at={(0.5,-0.22)},
            anchor=north,legend columns=2, font=\scriptsize},
        scaled ticks=base 10:-3,
        symbolic x coords={0.5\%, 1\%, 5\%, Full},
        xtick=data,
        ymajorgrids=true,
    ]

    \addplot coordinates {(0.5\%, 278) (5\%, 546) (1\%, 297) (Full, 0)};
    \addplot coordinates {(0.5\%, 99) (5\%, 208) (1\%, 115) (Full, 0)};
    \addplot coordinates {(0.5\%, 374) (5\%, 459) (1\%, 453) (Full, 10201)};

    \addlegendimage{red, line legend, solid, mark=square*}

    \node[above] at (axis cs:1\%,865) {{\small 14.4m}};
    \node[above] at (axis cs:5\%,1213) {{\small 20.2m}};
    \node[above] at (axis cs:0.5\%,751) {{\small 12.5m}};
    \node[above] at (axis cs:Full,10201) {{\small 170m}};
    \end{axis}

    \begin{axis}[
        axis y line*=right,
        axis x line=none, 
        ymin=0, ymax=120,
        ylabel={Performance (\%)},
        yticklabel style={align=right},
        width=0.32\linewidth,
        height=4.1cm,
        symbolic x coords={0.5\%, 1\%, 5\%, Full},
    ]

    \addplot[color=red, mark=square*] coordinates {(0.5\%, 91) (1\%, 93) (5\%, 97) (Full, 100)};
    \node[below] at (axis cs:5\%,97) {97\%};
    \node[below] at (axis cs:1\%,93) {93\%};
    \node[below] at (axis cs:0.5\%,91) {91\%};
    \end{axis}
    
\end{tikzpicture}}
    \vspace{-0.3cm}
    \caption{\textbf{Performance and runtime trade-off between our method and complete dataset training.} \textmd{
    The red line (``Performance'') represents the AUC ratio between the model trained on our result coresets and the original dataset. ``m'' stands for minutes.
    }}
    \label{fig:rq2-full}
\end{figure*}

\vspace{-0.2cm}
\subsection{Hyperparameters}
\label{sec:hyperparameters}
In each model, we employ a multi-layer perceptron with three layers (100-100-100) with ReLU activation and batch normalization. The dense embedding vector's hidden size is 10.
$N_{choose}$ is set to 3 for $0.5\%$ and $1\%$ data size, and 5 for $5\%$ data size. All missing features due to dataset reduction will be mapped to OOV tokens during final training. All methods are trained using the Adam optimizer \cite{kingma2014adam} with a learning rate of 0.01 and a batch size of 8192. We employ early stopping in our experiment, with the patience threshold set to one. Weight decay is searched in $\{ 10^{-7}, 5\times 10^{-7}, 10^{-6}, 5 \times 10^{-6},  10^{-5},  5\times10^{-5}, 10^{-4}, 5 \times 10^{-4}  \}$.

\section{Additional Experimental Results}
\label{sec:exp-extra}


\subsection{Further Efficiency Results (RQ2)}

Figure \ref{fig:rq2-full} shows the trade-off between performance, runtime, and coreset size of NaCS compared to original dataset training. The experiment results demonstrate that our method can significantly reduce the training time of content-based recommendation on various dataset, while maintaining similar performance, especially in larger datasets (e.g., KDD).

\begin{figure}[thb]
    \centering
    \pgfplotstableread[col sep=comma]{figures/rq6/data/criteo_remove_rate.csv}\criteo
\pgfplotstableread[col sep=comma]{figures/rq6/data/avazu_remove_rate.csv}\avazu

\begin{tikzpicture}
\begin{axis}[
    hide axis,
    xmin=0, xmax=1, ymin=0, ymax=1,
    legend cell align=left,
    legend style={
        draw=none,
        at={(0.5,0.5)},
        anchor=center,
        legend columns=2,
        font=\scriptsize
    }
]
\addlegendimage{mark=*, color=blue}
\addlegendentry{Criteo}

\addlegendimage{mark=square*, red}
\addlegendentry{Avazu}
\end{axis}
\end{tikzpicture}

\begin{tikzpicture}
\tikzstyle{every node}=[font=\scriptsize]
    \begin{axis}[
        width=0.5\linewidth,
        height=3cm,
        xlabel={Remove Ratio},
        ylabel={AUC},
        xtick={0,5,10,15,20}
    ]
    \addplot table[
        x=rate, 
        y=auc
    ]
    {\criteo};

    \addplot table[
        x=rate, 
        y=auc
    ]
    {\avazu};
    \end{axis}
    
\end{tikzpicture}
\hfill
\begin{tikzpicture}
\tikzstyle{every node}=[font=\scriptsize]
    \begin{axis}[
        width=0.5\linewidth,
        height=3cm,
        xlabel={Remove Ratio},
        ylabel={Log Loss},
        xtick={0,5,10,15,20}
    ]
    \addplot table[
        x=rate, 
        y=logloss
    ]
    {\criteo};

    \addplot table[
        x=rate, 
        y=logloss
    ]
    {\avazu};
    \end{axis}
    
\end{tikzpicture}
    \vspace{-0.3cm}
    \caption{Ablation study results for remove rate.}
    \label{fig:rq6-remove-rate}
\end{figure}
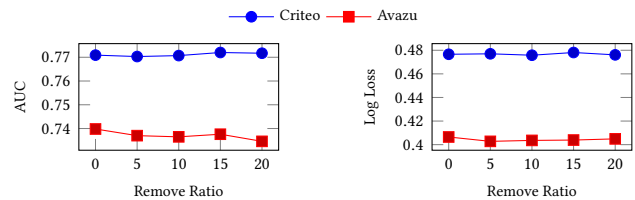

\subsection{Hyperparameter Sensitivity}
\label{sec:exp-hyperparam}

For this experiment, we employ DCNv2 as the backbone  to study how our proposed method interacts with various hyperparameters.

\noindent \textbf{Remove rate} In this experiment, we study how the amount of removed data affects the model performance. To reduce the required compute resource, we fix the noisy dataset size to 1.11\% (removing 10\% will result in 1\% coreset). Figure \ref{fig:rq6-remove-rate} shows the results. In the Criteo dataset, the performance remains stable through the denoise process. Interestingly, the performance slightly increases at a 15\% remove rate. 
In contrast, in the Avazu dataset, as we remove more samples, the performance drastically degrades, especially after 15\%.

\begin{figure}[thb]
    \centering
    \begin{subfigure}{0.48\textwidth}
    \centering
    \begin{tikzpicture}
\begin{axis}[
    hide axis,
    xmin=0, xmax=1, ymin=0, ymax=1,
    legend cell align=left,
    legend style={
        draw=none,
        at={(0.5,0.5)},
        anchor=center,
        legend columns=2,
        font=\scriptsize
    }
]
\addlegendimage{mark=*, color=blue}
\addlegendentry{Criteo}

\addlegendimage{mark=square*, red}
\addlegendentry{Avazu}
\end{axis}
\end{tikzpicture}
    \end{subfigure}
    
    \subcaptionbox{1\%\vspace{0.2cm}}{\pgfplotstableread[col sep=comma]{figures/rq6/data/criteo_n_choose.csv}\criteo
\pgfplotstableread[col sep=comma]{figures/rq6/data/avazu_n_choose.csv}\avazu

\begin{tikzpicture}
\tikzstyle{every node}=[font=\scriptsize]
    \begin{axis}[
        width=0.5\linewidth,
        height=3cm,
        xlabel={$N_{choose}$},
        ylabel={AUC},
        xtick={1,2,3,4,5}
    ]
    \addplot table[
        x=n_choose, 
        y=auc
    ]
    {\criteo};

    \addplot table[
        x=n_choose, 
        y=auc
    ]
    {\avazu};
    \end{axis}
    
\end{tikzpicture}
\hfill
\begin{tikzpicture}
\tikzstyle{every node}=[font=\scriptsize]
    \begin{axis}[
        width=0.5\linewidth,
        height=3cm,
        xlabel={$N_{choose}$},
        ylabel={Log Loss},
        xtick={1,2,3,4,5}
    ]
    \addplot table[
        x=n_choose, 
        y=logloss
    ]
    {\criteo};

    \addplot table[
        x=n_choose, 
        y=logloss
    ]
    {\avazu};
    \end{axis}
    
\end{tikzpicture}\vspace{-0.2cm}}

    \subcaptionbox{5\%}{\pgfplotstableread[col sep=comma]{figures/rq6/data/criteo_n_choose.csv}\criteo
\pgfplotstableread[col sep=comma]{figures/rq6/data/avazu_n_choose.csv}\avazu

\begin{tikzpicture}
\tikzstyle{every node}=[font=\scriptsize]
    \begin{axis}[
        width=0.5\linewidth,
        height=3cm,
        xlabel={$N_{choose}$},
        ylabel={AUC},
        xtick={1,2,3,4,5}
    ]
    \addplot table[
        x=n_choose, 
        y=auc5
    ]
    {\criteo};

    \addplot table[
        x=n_choose, 
        y=auc5
    ]
    {\avazu};
    \end{axis}
    
\end{tikzpicture}
\hfill
\begin{tikzpicture}
\tikzstyle{every node}=[font=\scriptsize]
    \begin{axis}[
        width=0.5\linewidth,
        height=3cm,
        xlabel={$N_{choose}$},
        ylabel={Log Loss},
        xtick={1,2,3,4,5}
    ]
    \addplot table[
        x=n_choose, 
        y=logloss5
    ]
    {\criteo};

    \addplot table[
        x=n_choose, 
        y=logloss5
    ]
    {\avazu};
    \end{axis}
    
\end{tikzpicture}\vspace{-0.2cm}}
    \vspace{-0.2cm}    
    \caption{Ablation study results for number of selection steps ($N_{choose}$) under various data size budgets.}
    \vspace{-0.3cm}
    \label{fig:rq6-n-choose}
\end{figure}
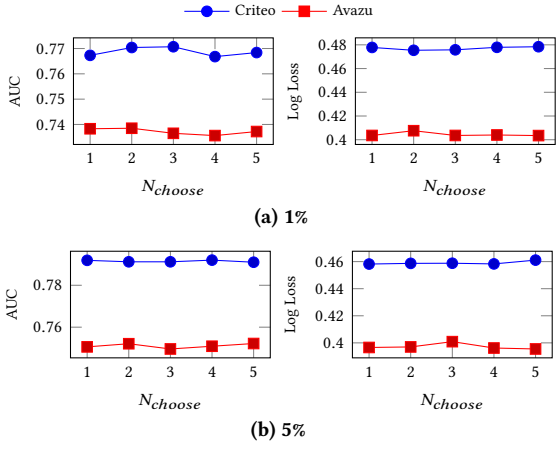

\noindent \textbf{Number of selection steps} In this experiment, we analyze how the number of selection steps affects the performance. As we want to verify that larger coresets require more selection steps, we conduct the experiment on both 1\% and 5\% coreset sizes. Figure \ref{fig:rq6-n-choose} shows the results. For both Criteo and Avazu datasets, increasing the number of coreset selection epochs ($N_{choose}$) generally improves performance up to a point, after which the gains diminish or slightly regress. 
For instance, at 1\% coreset size, the performance modestly improves up to 2-3 epochs. However, further steps yield little benefit or degrade performance, likely due to overfitting to noise. 
In contrast, at 5\% coreset size, the performance is more stable, indicating that larger coresets enable more robust label correction.
Moreover, larger coresets have a higher optimal $N_{choose}$, as more samples minimize the risk of overfitting.

\subsection{Learning Curve}
\label{sec:learn-curve}

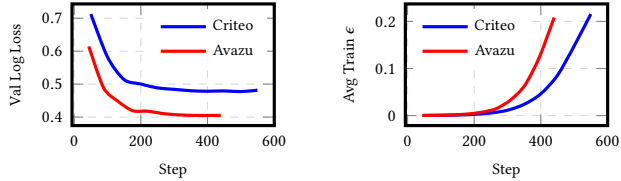
\begin{figure}[bth]
    \centering
    \vspace{-0.3cm}
    \subcaptionbox{Validation Loss}
   {

\pgfplotstableread[col sep=comma]{figures/rebuttal_rq1/logloss_criteo.csv}\criteo
\pgfplotstableread[col sep=comma]{figures/rebuttal_rq1/logloss_avazu.csv}\avazu

\begin{tikzpicture}
    \begin{axis}[
        width=0.24\textwidth,
        height=3.2cm,
        xlabel={Step},
        ylabel={Val Log Loss},
        grid=major,
        grid style={dashed,gray!30},
        legend pos=north east,
        line width=1.2pt,
        mark size=0pt,
        xlabel style={font=\scriptsize},
        legend style={font=\scriptsize},
        ylabel style={font=\scriptsize},
        tick label style={font=\scriptsize},
        legend style={font=\scriptsize, draw=none},
        legend entries={Criteo, Avazu},
    ]

    \addplot[blue, smooth] table[x=step, y=value] {\criteo};
    \addplot[red, smooth] table[x=step, y=value] {\avazu};


    \end{axis}
\end{tikzpicture}}
    \hfill
    \subcaptionbox{Train $\epsilon$}{

\pgfplotstableread[col sep=comma]{figures/rebuttal_rq1/train_eps_criteo.csv}\criteo
\pgfplotstableread[col sep=comma]{figures/rebuttal_rq1/train_eps_avazu.csv}\avazu

\begin{tikzpicture}
    \begin{axis}[
        width=0.24\textwidth,
        height=3.2cm,
        xlabel={Step},
        ylabel={Avg Train $\epsilon$},
        grid=major,
        grid style={dashed,gray!30},
        legend pos=north west,
        line width=1.2pt,
        mark size=0pt,
        xlabel style={font=\scriptsize},
        ylabel style={font=\scriptsize},
        tick label style={font=\scriptsize},
        legend style={font=\scriptsize},
        legend entries={Criteo, Avazu},
        legend style={font=\scriptsize, draw=none},
    ]

    \addplot[blue, smooth] table[x=step, y=value] {\criteo};
    \addplot[red, smooth] table[x=step, y=value] {\avazu};


    \end{axis}
\end{tikzpicture}}
    \vspace{-0.3cm}
    \caption{Visualization of the learning curve.}
    \label{fig:val-logloss}
\end{figure}

To study the effectiveness of the proposed progressive label self-correction, we analyze the validation log loss and the trust value $\epsilon$ on 1\% result subsets. As shown in  Fig. \ref{fig:val-logloss}, as the training progress, the validation log loss consistently decreases, showing that model predictions become increasingly reliable over time. Additionally, we employ early stopping based on validation performance to mitigate error accumulation during self-correction. 

\subsection{Effectiveness of Final Layer Gradient}
\label{sec:last-layer}

\begin{table}[h]
    \renewcommand{\arraystretch}{0.8}
    \setlength\tabcolsep{2.5pt}
    \centering
    \caption{Ablation study on the effect of gradient used for selecting coresets. \textmd{Runtime only includes selection time. In short, using the final layer gradient improves both efficiency and accuracy.}}
    \vspace{-0.2cm}
    \label{tab:n_last_layers}
    \begin{tabular}{lcccc}
    \toprule
        Dataset & Gradient Layer(s) & AUC & LogLoss & Runtime (s) \\
    \midrule
        \multirow{2}{*}{Criteo} & Final Layer & 0.7707 & 0.4758 & 165 \\
            & Last 3 Layers & 0.7711 & 0.4759 & 177 \\
        \multirow{2}{*}{Avazu} & Final Layer & 0.7365 & 0.4063 & 191 \\
         & Last 3 Layers & 0.7312 & 0.4182 & 208 \\
    \bottomrule
    \end{tabular}


\end{table}

We perform an ablation study on the effect of using only the gradient from final layer in Table~\ref{tab:n_last_layers}. The experiments are conducted with the DCNv2 backbone and a 1\% subset size. We observe that when using gradient from more layers (the last 3 layers are used), the gradient is sparser due to ReLU and Dropout, leading to more noisy signals for coreset selection. 

\begin{table}[thb]
    \caption{\textbf{Comparison with naive greedy algorithm implementation.} \textmd{The results are shown in seconds. $k$ is number of point to select, $\vert \mathcal{D} \vert$ is the dataset size. The second header of Submodlib denotes the similarity kernel implementation. ``-'' denotes OOM. ``B'' refers to our batch size. Best result in each line is marked in bold.}}
    \setlength\tabcolsep{2.8pt}
    \renewcommand{\arraystretch}{0.7}
    \vspace{-0.3cm}
    \begin{tabular}{lccccc}
        \toprule
        (k, $\vert \mathcal{D} \vert$)  & Deep & \multicolumn{2}{c}{Submodlib} & \multicolumn{2}{c}{Ours} \\
        \cmidrule(lr){3-4} \cmidrule(lr){5-6}
        & Core & CPU & GPU & B=1 & B=4 \\
        \midrule
        (50, $10^2$) & 0.0033 & \textbf{0.0029} & 0.0108 & 0.0249 & 0.0220 \\
        (50, 500) & 0.0662 & 0.0419 & 0.0368 & 0.0248 & \textbf{0.0221} \\
        (50, $10^3$) & 0.2858 & 0.1496 & 0.0946 & 0.0301 & \textbf{0.0234}\\
        (100, $10^3$) & 0.3670 & 0.2624 & 0.2141 & 0.0322 & \textbf{0.0252}\\ 
        (50, $10^4$)& - & 22.3749 & 15.8574 & 0.1761 & \textbf{0.0929} \\ 
        (100, $10^4$)& - & 37.6951 & 31.8456 & 0.1931 & \textbf{0.1762} \\ 
        \bottomrule
    \end{tabular}
    \label{tab:naive-greedy}
\end{table}

\subsection{Efficiency of Our Implementation}
\label{sec:naive-greedy-comp}

Table \ref{tab:naive-greedy} further compares the naive greedy algorithm implementation with ours, demonstrating efficiency gain from our version.

\end{document}